%% file: main.tex
\documentclass[conference]{IEEEtran}
\IEEEoverridecommandlockouts
\usepackage{cite}
\usepackage{amsmath,amssymb,amsfonts}
\usepackage{algorithmic}
\usepackage{graphicx}
\usepackage{textcomp}
\usepackage{xcolor}
\usepackage{url}
\usepackage{booktabs}
\usepackage{multirow}
\usepackage{graphicx}
\usepackage{svg}  
\def\BibTeX{{\rm B\kern-.05em{\sc i\kern-.025em b}\kern-.08em
    T\kern-.1667em\lower.7ex\hbox{E}\kern-.125emX}}
\begin{document}

\title{AURA: Agent for Understanding, Reasoning, and
Automated Tool Use in Voice-Driven Tasks
}

\author{
\IEEEauthorblockN{Leander Melroy Maben,
Gayathri Ganesh Lakshmy\textsuperscript{*},
Srijith Radhakrishnan\textsuperscript{*},
Siddhant Arora,
Shinji Watanabe
}\\
\IEEEauthorblockA{Carnegie Mellon University\\
\{lmaben, gganeshl, srijithr, siddhana, swatanab\}@cs.cmu.edu}
\thanks{\textsuperscript{*}Equal contribution.}
}


\maketitle

\begin{abstract}
Despite advances in language and speech technologies, no open-source system enables full speech-to-speech, multi-turn dialogue with integrated tool use and agentic reasoning. We introduce \textbf{AURA} (Agent for Understanding, Reasoning, and Automated Tool Use), the first open-source, speech-native assistant capable of completing complex, goal-driven tasks through dynamic tool invocation and multi-turn conversation. AURA combines open-weight ASR, TTS, and LLMs in a cascaded pipeline and supports tools such as calendar booking, contact lookup, web search, and email. Its modular design allows easy integration of new tools using natural language prompts and action classes. On VoiceBench, AURA scores \textbf{92.75\%} on OpenBookQA—outperforming all open-weight systems and nearing GPT-4o—and \textbf{4.39} on AlpacaEval, competitive with other open-weight systems. Human evaluation shows \textbf{90\%} task success on complex, multi-turn speech tasks.
\end{abstract}


\input{sections/introduction}

\input{sections/system}

\input{sections/experiments}

\input{sections/conclusion}

\bibliographystyle{IEEEtran}
\bibliography{main}

\end{document}

%% file: sections/introduction.tex
\section{Introduction}

\begin{figure*}[t]
\centering
\includegraphics[width=0.7\textwidth]{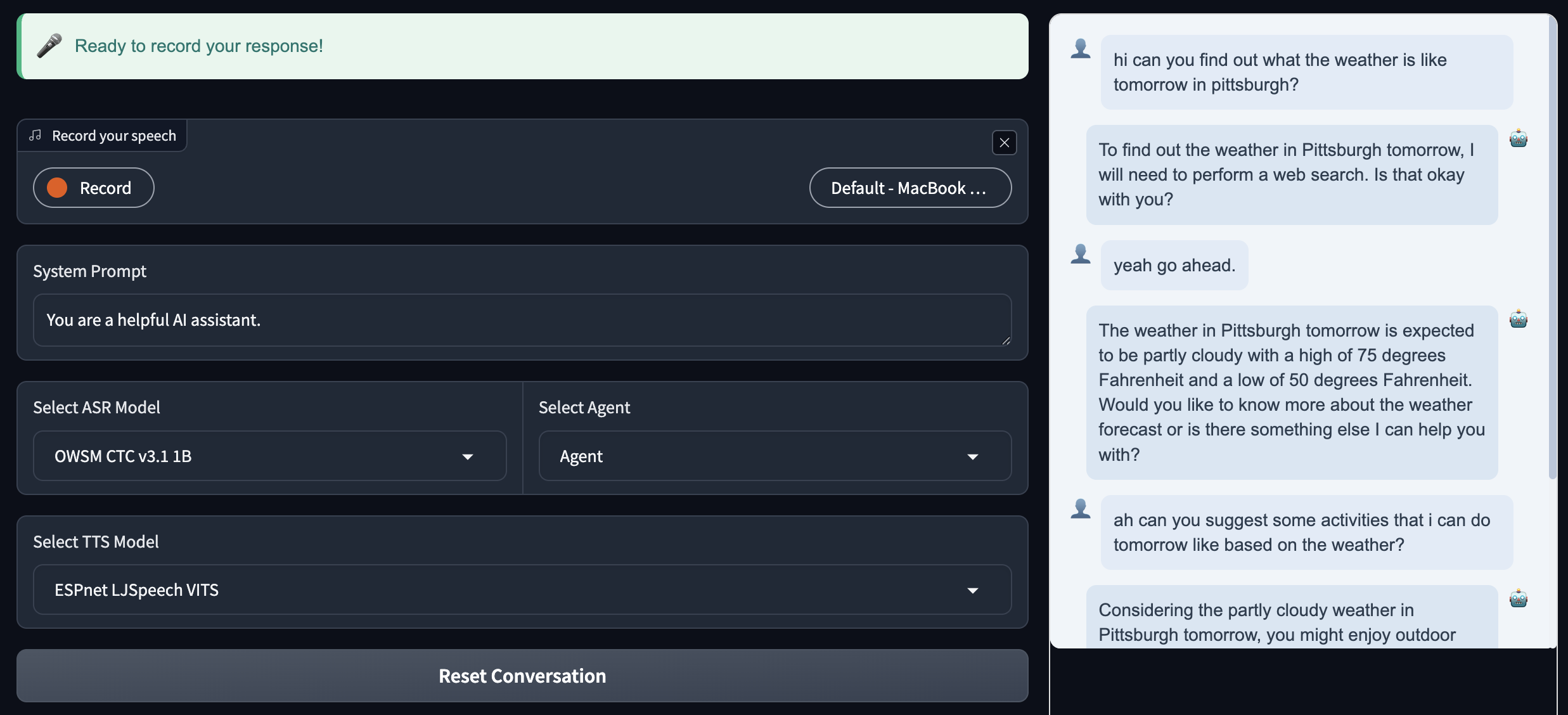}
\caption{User Interface for AURA}
\label{fig:ui}
\end{figure*}

Task-oriented conversational agents help users achieve real-world goals, such as booking travel, scheduling meetings, performing web searches, or sending emails, through natural language dialogue. These systems must interpret user utterances, maintain a dialogue state (structured or implicit), and generate appropriate actions or responses. A common approach uses a cascaded architecture with modular components for ASR, NLU/NLG, and TTS, as in ESPnet-SDS \cite{arora2025espnetsdsunifiedtoolkitdemo}. In contrast, end-to-end (E2E) models like SpeechGPT \cite{speechgpt}, Baichuan-Audio \cite{li2025baichuanaudiounifiedframeworkendtoend}, Moshi \cite{moshi}, Mini-Omni \cite{xie2024miniomnilanguagemodelshear} and Open-Omni \cite{luo2025openomniadvancingopensourceomnimodal} aim to generate speech directly from speech. However, benchmarks such as VoiceBench \cite{chen2024voicebenchbenchmarkingllmbasedvoice} indicate that end-to-end systems currently underperform cascaded pipelines on practical tasks, likely because high-performing text-based large language models are more readily available and easier to adapt within cascaded settings.

While large language models have been extensively employed for dialogue state tracking and response generation on text-based benchmarks such as MultiWOZ (Multi-Domain Wizard-of-Oz) \cite{budzianowski2018multiwoz}, their application to spoken dialogue remains relatively underexplored. The SpokenWOZ dataset \cite{si2024spokenwozlargescalespeechtextbenchmark} extends MultiWOZ to the speech domain using ASR transcripts, benchmarking dialogue state tracking and generation with closed-source models such as InstructGPT$_{003}$ and ChatGPT \cite{ouyang2022traininglanguagemodelsfollow}. However, these benchmarks focus narrowly on perception and surface-level dialogue capabilities and do not support the use of real-world tools, which is a crucial shortcoming that limits practical applicability. Without such capabilities, these systems would be unable to access dynamic, time-sensitive information (e.g., weather, geopolitical updates, alerts) or interact productively with everyday applications (e.g., calendar, email, contacts), severely limiting their effectiveness in real-world scenarios.

Several tool-augmented systems such as HuggingGPT \cite{shen2023hugginggptsolvingaitasks}, API-Bank \cite{li2023apibankcomprehensivebenchmarktoolaugmented}, ToolDial \cite{shim2025tooldialmultiturndialoguegeneration}, and AutoTOD \cite{xu-etal-2024-rethinking} have explored the use of tools in text-based systems. However, some lack the ability to perform compound tasks, where users issue follow-up requests based on prior results, and none support spoken interactions. This leaves a substantial gap in the development of speech-driven systems capable of both natural dialogue and real-world tool use.

We introduce \textbf{AURA}\footnote{
\begin{tabular}[t]{@{}l@{}}
Code: \url{https://github.com/Sentientia/Aura} \\
Demo: \url{https://www.youtube.com/watch?v=cb7w0GVwwF0}
\end{tabular}
}, 
the first \textbf{open-source}, \textbf{speech-to-speech} task-oriented agent that combines \textbf{reasoning} and \textbf{tool-use} capabilities. AURA employs a lightweight cascaded architecture and integrates a ReAct-style agent \cite{yao2023react} to interleave reasoning and action, a method shown to enhance multi-step task performance. Its design draws from recent advances in audio-based reasoning, such as Audio-CoT \cite{ma2025audiocotexploringchainofthoughtreasoning} and Audio Reasoner \cite{xie2025audioreasonerimprovingreasoningcapability}, and is inspired by dialog-driven agents such as ReSpAct \cite{dongre2025respactharmonizingreasoningspeaking}.

AURA supports a wide range of real-world \textbf{tool-use APIs}, including calendar booking, contact retrieval, email composition, and web search. It handles \textbf{multi-turn dialogue}, \textbf{follow-up requests}, and \textbf{compound goals}. On the VoiceBench benchmark \cite{chen2024voicebenchbenchmarkingllmbasedvoice}, AURA achieves \textbf{92.75\%} accuracy on OpenBookQA \cite{OpenBookQA2018}, surpassing all open-weight models and approaching the performance of the closed Whisper-v3-Large \cite{radford2022whisper} + GPT-4o \cite{openai2024gpt4o} system (92.97\%). In human evaluation of over 30 compound tasks, AURA achieves a \textbf{ 90\%} success rate, demonstrating strong capability in real-world task execution through speech. The user interface used for interaction is illustrated in Fig.~\ref{fig:ui}.

%% file: sections/system.tex
\section{System Design and Method}

\begin{figure}[t]
\centering
\includegraphics[width=0.43\textwidth]{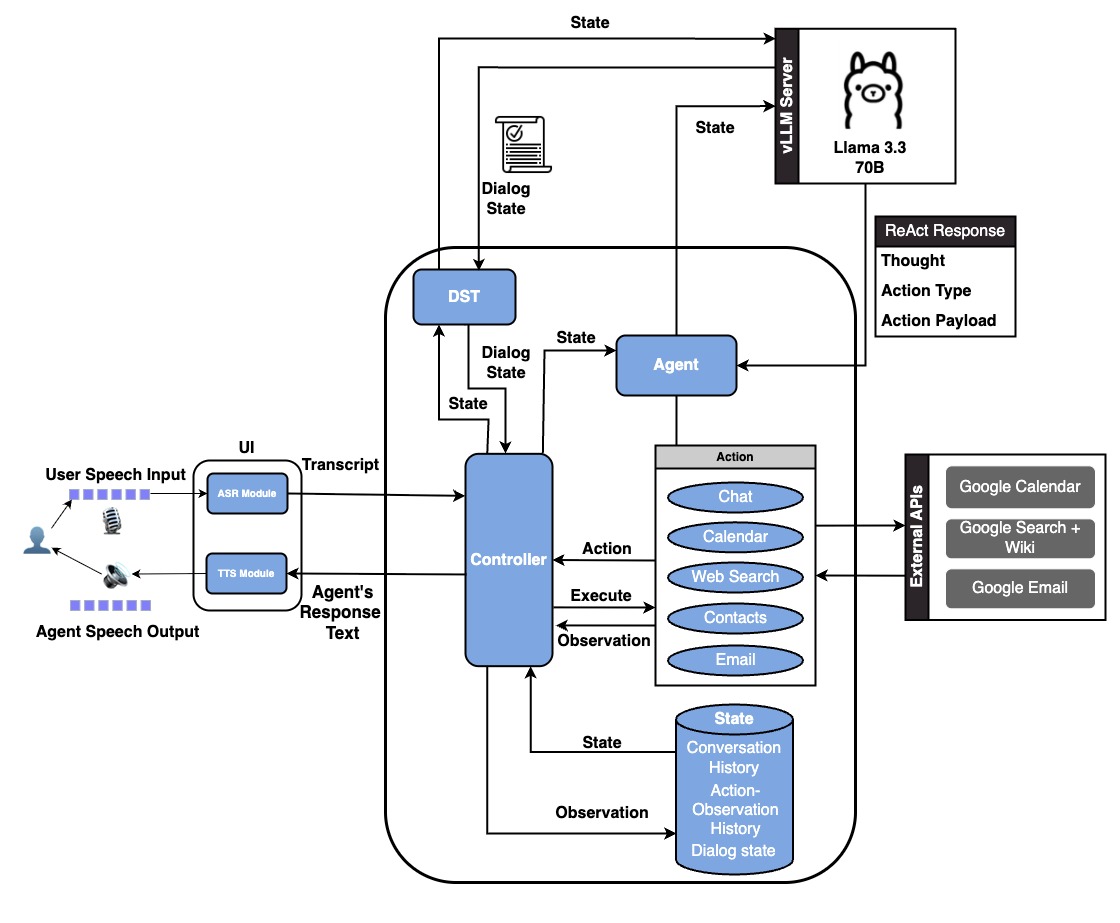}
\caption{Software Architecture for AURA}
\label{fig:system_block_diagram}
\end{figure}

We implement AURA using a cascaded architecture comprising four main components: \textbf{(i)} User Interface (UI), \textbf{(ii)} Dialog Processing Unit (DPU), \textbf{(iii)} LLM Server, and \textbf{(iv)} External APIs. Fig.~\ref{fig:system_block_diagram} illustrates the architecture of the system.

\subsection{User Interface}

The graphical user interface (GUI) is built using Gradio, which allows users to speak to the agent, listen to spoken responses, and view the live transcript of the conversation. It supports flexible, multi-turn interactions by enabling users to activate voice input at any point during the dialogue. The interface integrates an Automatic Speech Recognition (ASR) module powered by ESPnet’s OWSM models \cite{peng2024owsmv31betterfaster} and Whisper-v3-larger \cite{radford2022whisper} and a Text-to-Speech (TTS) module using the kan-bayashi/ljspeech\_vits model from ESPnet-TTS \cite{hayashi2020espnet}.

\subsection{Dialog Processing Unit}

\subsubsection{Controller}
The Controller serves as the central orchestrator, interfacing between the UI and the rest of the system. 

\subsubsection{Agent}
The agent follows the ReAct paradigm \cite{yao2023react}, interleaving reasoning and action. 

\subsubsection{Actions}
Actions define executable operations. Each action encapsulates a payload and an execute method. We implement five types:

\begin{itemize}
\item \textbf{Chat:} Delivers messages to the user—questions, confirmations, or responses.
\item \textbf{Calendar:} Books or Edits calendar events
\item \textbf{Web Search:} Queries the web and Wikipedia; returns top-3 results per source.
\item \textbf{Contact:} Retrieves recent email contacts
\item \textbf{Email:} Sends emails with fields. 
\end{itemize}

\subsubsection{Observation}
Observations are environmental feedback resulting from the execution of actions, e.g. user responses or the success / failure of email delivery.

\subsubsection{Dialog State Tracking}
This optional module tracks structured dialogue state information over multiple turns using prompt-based inference with a large language model (LLM).

\subsubsection{State}
The system state maintains:
\begin{itemize}
\item \textbf{Action-Observation History:} A sequence of action-observation pairs, starting with the initial user utterance.
\item \textbf{Conversation History:} A filtered version of history retaining only chat interactions for convenience of operations such as DST.
\item \textbf{Dialog State:} A structured JSON object that maintains key-value pairs representing salient information exchanged throughout the dialogue.

\end{itemize}

\subsection{LLM Server}

We use the instruction-tuned LLaMA-3.3-Instruct(70B) \cite{grattafiori2024llama3herdmodels} model to power both agent reasoning and DST.

\subsubsection{vLLM}
We host the model using vLLM \cite{vllms}, a memory-efficient inference engine that utilizes PagedAttention for high-throughput generation.

\subsubsection{ReAct Response Format}
The LLM response to the agent is structured as follows: \textbf{(i)} Thought – reasoning behind the decision, \textbf{(ii)} Action type, and \textbf{(iii)} Payload – the information required for execution.

\subsection{Workflow}
The dialogue flow begins when the user speaks through the Gradio interface. The ASR module transcribes the audio input, and the resulting text is passed to the Controller which then updates state. 
This updated state is then forwarded to the agent, which queries the LLM to determine the next appropriate action.

The Controller executes the action—this may involve interacting with an external API (e.g., sending an email, performing a web search, or booking a calendar event). Once the action is executed, an observation (e.g., system feedback or user response) is recorded, and the state is updated accordingly.

If relevant, the system triggers Dialog State Tracking (DST) and updates the state. Finally, the system generates a response, delivers it to the user through text-to-speech (TTS), and then waits for the next user input to continue the interaction in a multi-turn, agentic loop.

\subsection{Security and Privacy Safeguards}
Following initial authentication (via Google login for access to email, calendar, and contacts), an access token is generated and stored locally, never uploaded to any external server or database. To prevent unintended communication, the system enforces a configurable whitelist, ensuring emails are only sent to approved recipients.

%% file: sections/experiments.tex
\section{Experiments and Results}

\subsection{Spoken Question Answering with Agentic Web Search}

We evaluated AURA's reasoning and tool use capabilities in two VoiceBench QA tasks \cite{chen2024voicebenchbenchmarkingllmbasedvoice}: \textbf{AlpacaEval} \cite{alpaca_eval} (open-ended, scored 1–5 by GPT-4o-mini \cite{openai2024gpt4omini}) and \textbf{OpenBookQA} \cite{OpenBookQA2018} (multiple-choice with ground-truth answers).

Table~\ref{tab:voicebench_eval_grouped} compares AURA with both open- and closed-weight systems, using baselines drawn from the VoiceBench leaderboard. These include the top-performing model in each category, along with other representative systems. Using Whisper-v3-large \cite{radford2022whisper} for ASR and LLaMA3.3-70B as the base language model, AURA achieves an accuracy of \textbf{92.75\%} on OpenBookQA—competitive with GPT-4o \cite{openai2024gpt4o} and outperforming all other open-weight systems while relying entirely on open components. Notably, prompting the agent to perform at least one \textbf{Web search before answering} improves performance on OpenBookQA’s multiple-choice questions, raising accuracy from \textbf{87.69\%} to \textbf{89.23\%}.

\begin{table}[h]
\centering
\small
\caption{
\textbf{VoiceBench Evaluation.} Performance of AURA and baselines on two QA tasks. 
\textbf{C} = Closed-weight, \textbf{O} = Open-weight, \textbf{E2E} = End-to-End (no ASR).
}
\label{tab:voicebench_eval_grouped}
\vspace{0.5em}
\scalebox{0.58}{
\begin{tabular}{llllc}
\toprule
\textbf{Task} & \textbf{Type} & \textbf{System / Base Model} & \textbf{ASR} & \textbf{Score (1-5) / Accuracy} \\
\midrule

\multirow{7}{*}{OpenBookQA} 
  & \multirow{2}{*}{C} 
    & GPT-4o-Audio \cite{openai2024gpt4oaudio} & E2E & 89.23\% \\
  &  & \textbf{GPT-4o} \cite{openai2024gpt4o} & \textbf{Whisper-v3-large} & \textbf{92.97\%} \cite{radford2022whisper} \\  
  \cmidrule{2-5}
  & \multirow{4}{*}{O} 
    & Moshi \cite{kyutai2024moshi} & E2E & 25.93\% \\
  &  & Mini-Omni2 \cite{xie2024miniomni2opensourcegpt4ovision} & E2E & 26.59\% \\
  &  & Kimi-Audio \cite{kimiteam2025kimiaudiotechnicalreport} & E2E & 83.52\% \\
  &  & AURA - LLaMA3.3-70B (Ours) & ESPnet-OWSM & 89.23\% \\
  &  & \textbf{AURA - LLaMA3.3-70B (Ours)} & \textbf{Whisper-v3-large} \cite{radford2022whisper} & \textbf{92.75\%} \\

\midrule
\multirow{7}{*}{AlpacaEval} 
  & \multirow{2}{*}{C} 
    & GPT-4o-Audio & E2E & 4.78 \\
  &  & \textbf{GPT-4o} & \textbf{Whisper-v3-large} & \textbf{4.80} \\  
  \cmidrule{2-5}
  & \multirow{4}{*}{O} 
    & Moshi & E2E & 2.01 \\
  &  & Mini-Omni2  & E2E & 2.32 \\
  &  & AURA - LLaMA3.3-70B (Ours) & ESPnet-OWSM & 4.33 \\
  &  & AURA - LLaMA3.3-70B (Ours) & Whisper-v3-large \cite{radford2022whisper} & 4.39 \\
  &  & \textbf{Qwen3-8B} \cite{qwen3technicalreport} & \textbf{Parakeet-TDT-0.6b-V2} \cite{nvidia2024parakeet} & \textbf{4.68} \\
\bottomrule
\end{tabular}
}
\end{table}

\subsection{Multi-Turn Task Execution via Human Feedback}

We evaluated AURA on 30 real-world tasks grouped by difficulty—Easy, Medium, and Hard—with harder tasks involving compound goals requiring multi-turn reasoning, memory, and tool use. A human evaluator (co-author) rates each interaction on difficulty, success, and satisfaction (1–5 scale); task details and scores are available in the repository\footnote{\url{https://github.com/Sentientia/Aura}}. As shown in Table~\ref{tab:human_eval_multiturn}, AURA demonstrates high success rates across all difficulty levels, with satisfaction scores consistently above 4. The agent reliably selects appropriate tools (e.g., web search, calendar) and interprets their outputs effectively to drive the conversation forward, highlighting its robustness in complex, speech-driven tasks.

\begin{table}[h]
\centering
\small
\caption{
\textbf{Human Evaluation of Multi-Turn Task Execution.}
Success and satisfaction across task difficulty levels for AURA.
}
\label{tab:human_eval_multiturn}
\vspace{0.5em}
\scalebox{0.85}{
\begin{tabular}{lccc}
\toprule
\textbf{Difficulty} & \textbf{Number of tasks} & \textbf{Avg. Success} & \textbf{Avg. Satisfaction (1–5)} \\
\midrule
Easy & 6   & 100\%     & 4.67 \\
Medium & 12 & 91.67\%   & 4.83 \\
Hard & 12  & 83.33\%   & 4.00 \\
\bottomrule
\end{tabular}
}
\end{table}

\subsection{Dialog State Tracking (DST)}

We evaluate AURA's dialog state tracking (DST) on the SpokenWOZ benchmark \cite{si2024spokenwozlargescalespeechtextbenchmark} using \textbf{Joint Goal Accuracy (JGA)}, which measures the exact match between predicted and ground-truth dialog states across turns. Domain classification is performed via prompt-based inference with the LLM prior to dialog state extraction.

As shown in Table~\ref{tab:jga_comparison}, AURA outperforms the best baseline (SPACE + WavLMalign) \cite{si2024spokenwozlargescalespeechtextbenchmark} by over 3 points, achieving a JGA of \textbf{28.76} using LLaMA3.3-70B with prompt-based DST.

\begin{table}[h]
    \centering
    \renewcommand{\arraystretch}{1.1}
    \setlength{\tabcolsep}{6pt}
    \caption{
    \textbf{Dialog State Tracking on SpokenWOZ.}
    AURA surpasses prior benchmarks on JGA without DST fine-tuning.
    }
    \label{tab:jga_comparison}
    \begin{tabular}{lc}
        \toprule
        \textbf{Model} & \textbf{JGA (\%)} \\
        \midrule
        SPACE + WavLMalign \cite{si2024spokenwozlargescalespeechtextbenchmark} (Best Baseline) & 25.65 \\
        \textbf{AURA - LLaMA3.3-70B (Ours)} & \textbf{28.76} \\
        \bottomrule
    \end{tabular}
\end{table}

%% file: sections/conclusion.tex
\section{Conclusion}

We present \textbf{AURA}, the first open-source, \textbf{speech-to-speech} assistant for task-oriented dialogue with integrated \textbf{reasoning} and \textbf{tool use}. AURA combines a \textbf{ReAct-style agent} for multi-turn control with real-world \textbf{APIs} (e.g., calendar, web search, email) and supports fully \textbf{voice-based interaction}. Its \textbf{modular design} enables easy extension via natural language–described \textbf{action classes}.

%% file: main.bbl
\begin{thebibliography}{10}
\providecommand{\url}[1]{#1}
\csname url@samestyle\endcsname
\providecommand{\newblock}{\relax}
\providecommand{\bibinfo}[2]{#2}
\providecommand{\BIBentrySTDinterwordspacing}{\spaceskip=0pt\relax}
\providecommand{\BIBentryALTinterwordstretchfactor}{4}
\providecommand{\BIBentryALTinterwordspacing}{\spaceskip=\fontdimen2\font plus
\BIBentryALTinterwordstretchfactor\fontdimen3\font minus \fontdimen4\font\relax}
\providecommand{\BIBforeignlanguage}[2]{{%
\expandafter\ifx\csname l@#1\endcsname\relax
\typeout{** WARNING: IEEEtran.bst: No hyphenation pattern has been}%
\typeout{** loaded for the language `#1'. Using the pattern for}%
\typeout{** the default language instead.}%
\else
\language=\csname l@#1\endcsname
\fi
#2}}
\providecommand{\BIBdecl}{\relax}
\BIBdecl

\bibitem{arora2025espnetsdsunifiedtoolkitdemo}
\BIBentryALTinterwordspacing
S.~Arora, Y.~Peng, J.~Shi, J.~Tian, W.~Chen, S.~Bharadwaj, H.~Futami, Y.~Kashiwagi, E.~Tsunoo, S.~Shimizu, V.~Srivastav, and S.~Watanabe, ``Espnet-sds: Unified toolkit and demo for spoken dialogue systems,'' 2025. [Online]. Available: \url{https://arxiv.org/abs/2503.08533}
\BIBentrySTDinterwordspacing

\bibitem{speechgpt}
\BIBentryALTinterwordspacing
D.~Zhang, S.~Li, X.~Zhang, J.~Zhan, P.~Wang, Y.~Zhou, and X.~Qiu, ``Speechgpt: Empowering large language models with intrinsic cross-modal conversational abilities,'' 2023. [Online]. Available: \url{https://arxiv.org/abs/2305.11000}
\BIBentrySTDinterwordspacing

\bibitem{li2025baichuanaudiounifiedframeworkendtoend}
\BIBentryALTinterwordspacing
T.~Li, J.~Liu, T.~Zhang, Y.~Fang, D.~Pan, M.~Wang, Z.~Liang, Z.~Li, M.~Lin, G.~Dong, J.~Xu, H.~Sun, Z.~Zhou, and W.~Chen, ``Baichuan-audio: A unified framework for end-to-end speech interaction,'' 2025. [Online]. Available: \url{https://arxiv.org/abs/2502.17239}
\BIBentrySTDinterwordspacing

\bibitem{moshi}
\BIBentryALTinterwordspacing
A.~Défossez, L.~Mazaré, M.~Orsini, A.~Royer, P.~Pérez, H.~Jégou, E.~Grave, and N.~Zeghidour, ``Moshi: a speech-text foundation model for real-time dialogue,'' 2024. [Online]. Available: \url{https://arxiv.org/abs/2410.00037}
\BIBentrySTDinterwordspacing

\bibitem{xie2024miniomnilanguagemodelshear}
\BIBentryALTinterwordspacing
Z.~Xie and C.~Wu, ``Mini-omni: Language models can hear, talk while thinking in streaming,'' 2024. [Online]. Available: \url{https://arxiv.org/abs/2408.16725}
\BIBentrySTDinterwordspacing

\bibitem{luo2025openomniadvancingopensourceomnimodal}
\BIBentryALTinterwordspacing
R.~Luo, T.-E. Lin, H.~Zhang, Y.~Wu, X.~Liu, M.~Yang, Y.~Li, L.~Chen, J.~Li, L.~Zhang, Y.~Chen, X.~Xia, H.~Alinejad-Rokny, and F.~Huang, ``Openomni: Advancing open-source omnimodal large language models with progressive multimodal alignment and real-time self-aware emotional speech synthesis,'' 2025. [Online]. Available: \url{https://arxiv.org/abs/2501.04561}
\BIBentrySTDinterwordspacing

\bibitem{chen2024voicebenchbenchmarkingllmbasedvoice}
\BIBentryALTinterwordspacing
Y.~Chen, X.~Yue, C.~Zhang, X.~Gao, R.~T. Tan, and H.~Li, ``Voicebench: Benchmarking llm-based voice assistants,'' 2024. [Online]. Available: \url{https://arxiv.org/abs/2410.17196}
\BIBentrySTDinterwordspacing

\bibitem{budzianowski2018multiwoz}
P.~Budzianowski, T.-H. Wen, B.-H. Tseng, I.~Casanueva, S.~Ultes, O.~Ramadan, and M.~Ga{\v{s}}i{\'c}, ``Multiwoz--a large-scale multi-domain wizard-of-oz dataset for task-oriented dialogue modelling,'' \emph{arXiv preprint arXiv:1810.00278}, 2018.

\bibitem{si2024spokenwozlargescalespeechtextbenchmark}
\BIBentryALTinterwordspacing
S.~Si, W.~Ma, H.~Gao, Y.~Wu, T.-E. Lin, Y.~Dai, H.~Li, R.~Yan, F.~Huang, and Y.~Li, ``Spokenwoz: A large-scale speech-text benchmark for spoken task-oriented dialogue agents,'' 2024. [Online]. Available: \url{https://arxiv.org/abs/2305.13040}
\BIBentrySTDinterwordspacing

\bibitem{ouyang2022traininglanguagemodelsfollow}
\BIBentryALTinterwordspacing
L.~Ouyang, J.~Wu, X.~Jiang, D.~Almeida, C.~L. Wainwright, P.~Mishkin, C.~Zhang, S.~Agarwal, K.~Slama, A.~Ray, J.~Schulman, J.~Hilton, F.~Kelton, L.~Miller, M.~Simens, A.~Askell, P.~Welinder, P.~Christiano, J.~Leike, and R.~Lowe, ``Training language models to follow instructions with human feedback,'' 2022. [Online]. Available: \url{https://arxiv.org/abs/2203.02155}
\BIBentrySTDinterwordspacing

\bibitem{shen2023hugginggptsolvingaitasks}
\BIBentryALTinterwordspacing
Y.~Shen, K.~Song, X.~Tan, D.~Li, W.~Lu, and Y.~Zhuang, ``Hugginggpt: Solving ai tasks with chatgpt and its friends in hugging face,'' 2023. [Online]. Available: \url{https://arxiv.org/abs/2303.17580}
\BIBentrySTDinterwordspacing

\bibitem{li2023apibankcomprehensivebenchmarktoolaugmented}
\BIBentryALTinterwordspacing
M.~Li, Y.~Zhao, B.~Yu, F.~Song, H.~Li, H.~Yu, Z.~Li, F.~Huang, and Y.~Li, ``Api-bank: A comprehensive benchmark for tool-augmented llms,'' 2023. [Online]. Available: \url{https://arxiv.org/abs/2304.08244}
\BIBentrySTDinterwordspacing

\bibitem{shim2025tooldialmultiturndialoguegeneration}
\BIBentryALTinterwordspacing
J.~Shim, G.~Seo, C.~Lim, and Y.~Jo, ``Tooldial: Multi-turn dialogue generation method for tool-augmented language models,'' 2025. [Online]. Available: \url{https://arxiv.org/abs/2503.00564}
\BIBentrySTDinterwordspacing

\bibitem{xu-etal-2024-rethinking}
\BIBentryALTinterwordspacing
H.-D. Xu, X.-L. Mao, P.~Yang, F.~Sun, and H.~Huang, ``Rethinking task-oriented dialogue systems: From complex modularity to zero-shot autonomous agent,'' in \emph{Proceedings of the 62nd Annual Meeting of the Association for Computational Linguistics (Volume 1: Long Papers)}, L.-W. Ku, A.~Martins, and V.~Srikumar, Eds.\hskip 1em plus 0.5em minus 0.4em\relax Bangkok, Thailand: Association for Computational Linguistics, Aug. 2024, pp. 2748--2763. [Online]. Available: \url{https://aclanthology.org/2024.acl-long.152/}
\BIBentrySTDinterwordspacing

\bibitem{yao2023react}
S.~Yao, J.~Zhao, D.~Yu, N.~Du, I.~Shafran, K.~Narasimhan, and Y.~Cao, ``React: Synergizing reasoning and acting in language models,'' in \emph{International Conference on Learning Representations (ICLR)}, 2023.

\bibitem{ma2025audiocotexploringchainofthoughtreasoning}
\BIBentryALTinterwordspacing
Z.~Ma, Z.~Chen, Y.~Wang, E.~S. Chng, and X.~Chen, ``Audio-cot: Exploring chain-of-thought reasoning in large audio language model,'' 2025. [Online]. Available: \url{https://arxiv.org/abs/2501.07246}
\BIBentrySTDinterwordspacing

\bibitem{xie2025audioreasonerimprovingreasoningcapability}
\BIBentryALTinterwordspacing
Z.~Xie, M.~Lin, Z.~Liu, P.~Wu, S.~Yan, and C.~Miao, ``Audio-reasoner: Improving reasoning capability in large audio language models,'' 2025. [Online]. Available: \url{https://arxiv.org/abs/2503.02318}
\BIBentrySTDinterwordspacing

\bibitem{dongre2025respactharmonizingreasoningspeaking}
\BIBentryALTinterwordspacing
V.~Dongre, X.~Yang, E.~C. Acikgoz, S.~Dey, G.~Tur, and D.~Hakkani-Tür, ``Respact: Harmonizing reasoning, speaking, and acting towards building large language model-based conversational ai agents,'' 2025. [Online]. Available: \url{https://arxiv.org/abs/2411.00927}
\BIBentrySTDinterwordspacing

\bibitem{OpenBookQA2018}
T.~Mihaylov, P.~Clark, T.~Khot, and A.~Sabharwal, ``Can a suit of armor conduct electricity? a new dataset for open book question answering,'' in \emph{EMNLP}, 2018.

\bibitem{radford2022whisper}
\BIBentryALTinterwordspacing
A.~Radford, J.~W. Kim, T.~Xu, G.~Brockman, C.~McLeavey, and I.~Sutskever, ``Robust speech recognition via large-scale weak supervision,'' 2022. [Online]. Available: \url{https://arxiv.org/abs/2212.04356}
\BIBentrySTDinterwordspacing

\bibitem{openai2024gpt4o}
OpenAI, ``Gpt-4o: Openai’s new multimodal flagship model,'' \url{https://platform.openai.com/docs/models/gpt-4o}, 2024, accessed: YYYY-MM-DD.

\bibitem{peng2024owsmv31betterfaster}
\BIBentryALTinterwordspacing
Y.~Peng, J.~Tian, W.~Chen, S.~Arora, B.~Yan, Y.~Sudo, M.~Shakeel, K.~Choi, J.~Shi, X.~Chang, J.~weon Jung, and S.~Watanabe, ``Owsm v3.1: Better and faster open whisper-style speech models based on e-branchformer,'' 2024. [Online]. Available: \url{https://arxiv.org/abs/2401.16658}
\BIBentrySTDinterwordspacing

\bibitem{hayashi2020espnet}
T.~Hayashi, R.~Yamamoto, K.~Inoue, T.~Yoshimura, S.~Watanabe, T.~Toda, K.~Takeda, Y.~Zhang, and X.~Tan, ``{Espnet-TTS}: Unified, reproducible, and integratable open source end-to-end text-to-speech toolkit,'' in \emph{Proceedings of IEEE International Conference on Acoustics, Speech and Signal Processing (ICASSP)}.\hskip 1em plus 0.5em minus 0.4em\relax IEEE, 2020, pp. 7654--7658.

\bibitem{grattafiori2024llama3herdmodels}
\BIBentryALTinterwordspacing
A.~G. et~al., ``The llama 3 herd of models,'' 2024. [Online]. Available: \url{https://arxiv.org/abs/2407.21783}
\BIBentrySTDinterwordspacing

\bibitem{vllms}
\BIBentryALTinterwordspacing
W.~Kwon, Z.~Li, S.~Zhuang, Y.~Sheng, L.~Zheng, C.~H. Yu, J.~E. Gonzalez, H.~Zhang, and I.~Stoica, ``Efficient memory management for large language model serving with pagedattention,'' 2023. [Online]. Available: \url{https://arxiv.org/abs/2309.06180}
\BIBentrySTDinterwordspacing

\bibitem{alpaca_eval}
X.~Li, T.~Zhang, Y.~Dubois, R.~Taori, I.~Gulrajani, C.~Guestrin, P.~Liang, and T.~B. Hashimoto, ``Alpacaeval: An automatic evaluator of instruction-following models,'' \url{https://github.com/tatsu-lab/alpaca_eval}, 5 2023.

\bibitem{openai2024gpt4omini}
OpenAI, ``Gpt-4o: Openai’s new multimodal flagship model,'' \url{https://openai.com/index/gpt-4o-mini-advancing-cost-efficient-intelligence/}, 2024, accessed: YYYY-MM-DD.

\bibitem{openai2024gpt4oaudio}
------, ``Gpt-4o: Openai’s new multimodal flagship model,'' \url{https://platform.openai.com/docs/models/gpt-4o-audio-preview}, 2024, accessed: YYYY-MM-DD.

\bibitem{kyutai2024moshi}
\BIBentryALTinterwordspacing
A.~D\'efossez, L.~Mazar\'e, M.~Orsini, A.~Royer, P.~P\'erez, H.~J\'egou, E.~Grave, and N.~Zeghidour, ``Moshi: a speech-text foundation model for real-time dialogue,'' Tech. Rep., 2024. [Online]. Available: \url{https://arxiv.org/abs/2410.00037}
\BIBentrySTDinterwordspacing

\bibitem{xie2024miniomni2opensourcegpt4ovision}
\BIBentryALTinterwordspacing
Z.~Xie and C.~Wu, ``Mini-omni2: Towards open-source gpt-4o with vision, speech and duplex capabilities,'' 2024. [Online]. Available: \url{https://arxiv.org/abs/2410.11190}
\BIBentrySTDinterwordspacing

\bibitem{kimiteam2025kimiaudiotechnicalreport}
\BIBentryALTinterwordspacing
KimiTeam, D.~Ding, Z.~Ju, Y.~Leng, S.~Liu, T.~Liu, Z.~Shang, K.~Shen, W.~Song, X.~Tan, H.~Tang, Z.~Wang, C.~Wei, Y.~Xin, X.~Xu, J.~Yu, Y.~Zhang, X.~Zhou, Y.~Charles, J.~Chen, Y.~Chen, Y.~Du, W.~He, Z.~Hu, G.~Lai, Q.~Li, Y.~Liu, W.~Sun, J.~Wang, Y.~Wang, Y.~Wu, Y.~Wu, D.~Yang, H.~Yang, Y.~Yang, Z.~Yang, A.~Yin, R.~Yuan, Y.~Zhang, and Z.~Zhou, ``Kimi-audio technical report,'' 2025. [Online]. Available: \url{https://arxiv.org/abs/2504.18425}
\BIBentrySTDinterwordspacing

\bibitem{qwen3technicalreport}
\BIBentryALTinterwordspacing
Q.~Team, ``Qwen3 technical report,'' 2025. [Online]. Available: \url{https://arxiv.org/abs/2505.09388}
\BIBentrySTDinterwordspacing

\bibitem{nvidia2024parakeet}
NVIDIA, ``Parakeet-tdt-0.6b-v2,'' \url{https://huggingface.co/nvidia/parakeet-tdt-0.6b-v2}, 2024, accessed: 2025-06-24.

\end{thebibliography}
